\documentclass[journal]{IEEEtran}
\usepackage{cite}

\usepackage{picinpar}

\ifCLASSINFOpdf
  \usepackage[pdftex]{graphicx}
\else
  \usepackage[dvips]{graphicx}
\fi
\usepackage[cmex10]{amsmath}
\usepackage{array}

\usepackage{ulem}
\usepackage{amsmath}
\usepackage{amssymb}
\usepackage{flushend}
\usepackage{color}
\usepackage{enumerate}
\usepackage{pbox}
\usepackage{multirow}
\usepackage{siunitx}

\begin{document}
%
\title{Robust Translational Force Control of Multi-Rotor UAV for Precise Acceleration Tracking*}
%
%
%

\author{Seung~Jae~Lee,~\IEEEmembership{Student Member,~IEEE,}
        Seung~Hyun~Kim,~\IEEEmembership{Member,~IEEE}
        and~H.~Jin~Kim,~\IEEEmembership{Member,~IEEE}
\thanks{*This work was supported by the Robotics Core Technology Development Project (10080301) funded by the MoTIE, Korea, and the MSIT (Ministry of Science and ICT), Korea, under the ITRC (Information Technology Research Center) support program (IITP-2019-2017-0-01637) supervised by the IITP (Institute for Information \& communications Technology Promotion).

Seung Jae Lee, Seung hyun Kim and Hyoun Jin Kim are with the Department
of Mechanical and Aerospace Engineering, School of Engineering, Seoul National University, Seoul, 08826 Republic of Korea. e-mail: sjlazza@snu.ac.kr, kshipme@snu.ac.kr, hjinkim@snu.ac.kr.}
}

\maketitle

\begin{abstract}
In this paper, we introduce a translational force control method with  disturbance observer (DOB)-based force disturbance cancellation  for precise three-dimensional acceleration control of a multi-rotor UAV.
The acceleration control of the multi-rotor requires conversion of the desired acceleration signal to the desired roll, pitch, and total thrust.
But because the attitude dynamics and the thrust dynamics are different, simple kinematic signal conversion without consideration of those difference can cause serious performance degradation in acceleration tracking.
Unlike most existing translational force control techniques that are based on such {simple} inversion, our new method allows controlling the acceleration of the multi-rotor more precisely by considering the dynamics of the multi-rotor during the kinematic inversion. 
By combining the DOB with the translational force system that includes the improved conversion technique, we achieve robustness with respect to the external force disturbances that hinders the accurate acceleration control. 
$\mu$-analysis is performed to ensure the robust  stability of the overall closed-loop system, considering the combined effect of various possible model uncertainties. 
Both simulation and experiment are conducted to validate the proposed technique, which confirms the satisfactory performance to track the desired acceleration of the multi-rotor.
\end{abstract}

Note to Practitioners:
\begin{abstract}
This paper presents a method for controlling the acceleration of a multi-rotor accurately under the presence of translational force disturbance.
Unlike the existing methods, the new signal conversion technique that considers the dynamics of the multi-rotor in the process of converting the target translational acceleration signal to the target roll, pitch and thrust signal enables a more accurate translational force control.
The disturbance observer (DOB) structure applied to the translational force control system overcomes the acceleration control performance deterioration caused by external translational force disturbance.
Through the combination of the two techniques, the acceleration of the multi-rotor can be accurately controlled not only in the nominal environment but also in the presence of translational force disturbance.
\end{abstract}

\begin{IEEEkeywords}
disturbance observer, $\mu$-analysis, multi-rotor, robust control, translational force control.
\end{IEEEkeywords}

%
\IEEEpeerreviewmaketitle

\IEEEPARstart{P}{recise} acceleration tracking is a fundamental requirement of multi-rotor unmanned aerial vehicles (UAVs) for widening their application area beyond basic autonomous flight. 
For such an objective, we need an accurate three-dimensional force control and a robust rejection method of translational motion disturbance. 


First, for accurate force control, the target force command must be converted to the appropriate target attitude and thrust value, because the multi-rotor generates the three-dimensional translational forces by the combination of the current attitude and the total propeller thrust \cite{mUAVDynamics}.
Once the target attitude and the total thrust command are determined, each value passes through attitude and thrust dynamics that are quite different from each other :  the process of achieving actual attitude involves feedback attitude control\cite{mofid2018adaptive}, torque generation by the combination of motor's thrust, followed by the rotation of the fuselage that has larger moment of inertia than the propellers.   
%
{Due to such difference, simple kinematic conversion of the force signal  without consideration of the actual attitude and thrust dynamics can cause unsynchronized realization of the attitude and the total thrust, which degrades acceleration tracking performance given that the acceleration of the multirotor is determined by the combination of attitude and thrust.}
To the best of our knowledge, however, many studies have not investigated this issue.
In  \cite{mUAVDynamics} and \cite{zuo2010trajectory}, the target thrust signal was computed without considering attitudinal dynamics while treating Z-directional translational dynamics as a separate channel to other horizontal dynamics.
All three axes of translational dynamics have been simultaneously considered in \cite{zhao2015nonlinear} during the conversion process, but they also did not reflect the different characteristics of attitude and thrust dynamics.
Those differences become noticeable in multi-rotors that have large moment of inertia, due to significant time delay between input and output attitude. 

\begin{figure*}[t]
\begin{center}
\includegraphics[width=18cm]{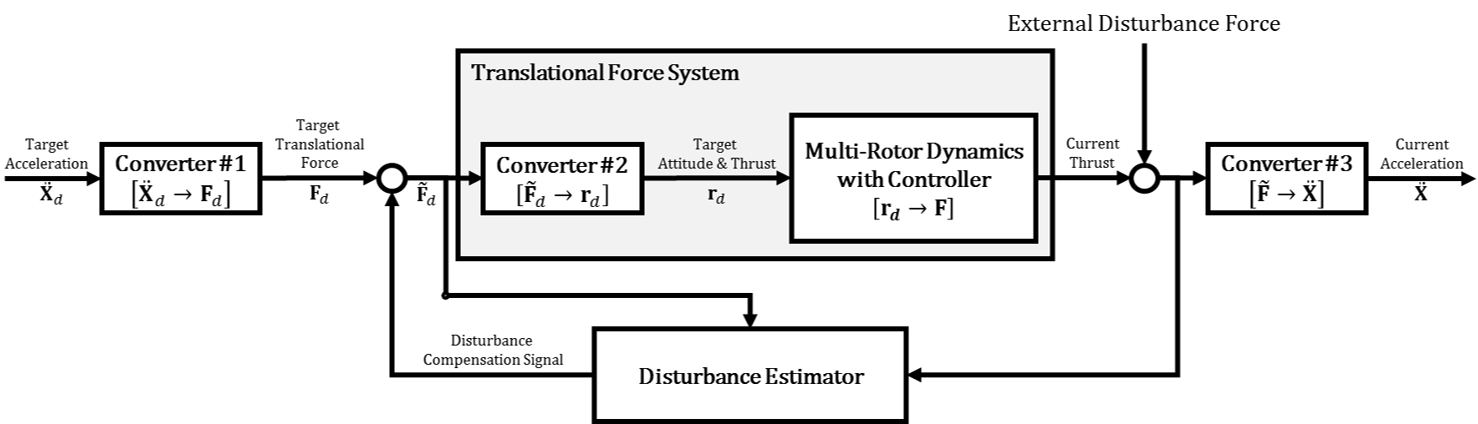}
\end{center}
\vspace{-0.5cm}
\caption{{Structure of the proposed translational force system with disturbance observer for precise and robust acceleration tracking performance of a multi-rotor UAV.}}
\label{gp:Intro_fig}
\end{figure*}

Second, for a satisfactory level of translational disturbance rejection, we need a controller that estimates and offsets the effect of the disturbance\cite{mobayen2016lmi, mobayen2018composite}.   
As a way to achieve this goal, we can consider constructing the Disturbance Observer (DOB)-based robust control algorithm \cite{ohnishi1996motion}.
However, although several studies applied the Disturbance Observer (DOB) robust control technique to their controllers \cite{Attitude_DOB_1,Attitude_DOB_2,Attitude_DOB_3,Attitude_DOB_4,Attitude_DOB_5,General_DOB,neural_DOB,suseong_DOB, DOB_star_1, DOB_star_2, DOB_sjlazza}, most of them  \cite{Attitude_DOB_1,Attitude_DOB_2,Attitude_DOB_3,Attitude_DOB_4,Attitude_DOB_5,General_DOB,neural_DOB,suseong_DOB} aimed to maintain the nominal attitude control performance against \textit{torque} disturbance.
Therefore, this approach has a limitation in overcoming translational movement disturbances.
Only a few studies exist on applying DOB to overcome the translational force disturbances including \cite{DOB_star_1}, \cite{DOB_star_2}, and \cite{DOB_sjlazza}. 
In \cite{DOB_star_1}, however, only the estimation method of the disturbance is introduced and no specific control method for overcoming the disturbance using the estimated disturbance is proposed.
In \cite{DOB_star_2}, inverse kinematics rather than inverse dynamics is used in the process of generating disturbance compensation signal. 
This approach can cause severe degradation in disturbance estimation performance as the dynamics is not negligible. 
In \cite{DOB_sjlazza}, which is the preliminary research of this paper, the structure of DOB to cope with translational force disturbance is proposed.
However, accurate translational acceleration control is not achieved because of the error in converting the target translational force command to the target attitude and total thrust.
Also, the nominal model used in DOB is  based on inaccurate desired acceleration-to-desired states conversion technique.

In this paper, we present a new accurate three-dimensional translational acceleration tracking control  that overcomes the limitations of the previous studies.
The contributions of the proposed acceleration control technique are as follows.
First, we introduce a new conversion method that reflects the difference between attitude dynamics and thrust dynamics when computing  the target attitude and total thrust command from the translational force command  {(i.e., `Converter \#2' block of Fig. \ref{gp:Intro_fig})}.
Second, we model the  translational force system (i.e., the shaded part of Fig. \ref{gp:Intro_fig}) that includes the new command conversion method, and design a DOB-based robust controller (i.e., `Disturbance Estimator' block of Fig. \ref{gp:Intro_fig}) that overcomes translational force disturbance based on our new model.
In the DOB controller design process, we perform $\mu$-analysis to systematically take into account the complex effects of various uncertainty.
By presenting simulation and experimental results, we demonstrate the target acceleration tracking performance of the proposed conversion technique and the ability to overcome the external translational force disturbance of the designed DOB controller.

This paper is organized as follows. In Section II, we discuss the mathematical model of the multi-rotor used in the controller design. Section III deals with the force control of multi-rotor, and Section IV describes how DOB is applied to the force control. Section V provides the stability analysis to determine the range of DOB parameters that guarantee the stability of the designed system even in the presence of various uncertain elements. In Section VI, we demonstrate the empirical validity through simulations and actual experiments.

\section{MODELLING OF MULTI-ROTOR UAV}

The rigid body dynamics of the multi-rotor are given by
\begin{equation}
\label{eq:dyn_full}
\left\{
\begin{array}{lr}
m\ddot{\mathbf{X}}=\mathbf{F}+m\mathbf{g}=R(\mathbf{q})\mathbf{T}_t+m\mathbf{g}\\
J\dot{\mathbf{\Omega}}=\mathbf{T}_r-\mathbf{\Omega}\times J\mathbf{\Omega} \;,
\end{array}
\right.
\end{equation}
where $m$ is the mass of the multi-rotor, $\mathbf{X}=[x\ y\ z]^T\in \mathbb{R}^{3\times 1}$ is the position in the earth fixed frame, $\mathbf{F}=[F_x\ F_y\ F_z]^T\in\mathbb{R}^{3\times 1}$ is the three-dimensional translational force vector generated by the multi-rotor, $R(\mathbf{q})$ is the rotation matrix from the body frame to earth fixed frame, $\mathbf{q}=[\phi\ \theta\ \psi]^T\in\mathbb{R}^{3\times 1}$ is an attitude of the multi-rotor in the earth fixed frame, $\mathbf{T}_t=[0\ 0\ -T_t]^T\in\mathbb{R}^{3\times 1}$ is the thrust force vector in the body frame, $T_t\in\mathbb{R}$ is the magnitude of the total thrust, and $\mathbf{g}=[0\ 0\ g]^T\in\mathbb{R}^{3\times1}$ is a gravity vector. The parameter $J\in\mathbb{R}^{3\times3}$ is the moment of inertia (MOI) of the multi-rotor, $\mathbf{\Omega}=[p\ q\ r]^T\in\mathbb{R}^{3\times1}$ is an angular velocity vector defined in the body frame, and $\mathbf{T}_r=[\tau_r\ \tau_p\ \tau_y]^T\in\mathbb{R}^{3\times 1}$ is an attitude control torque vector. For attitude dynamics, simplified dynamics of
\begin{equation}
\label{eq:dyn_rot_simplify}
J\ddot{\mathbf{q}}=\mathbf{T}_r
\end{equation}
is commonly used, taking into account the small operation range of roll and pitch angle of multi-rotor and negligible Coriolis term \cite{mUAVDynamics,DOB_star_2,proof_of_small_gyroscopic}.

\section{TRANSLATIONAL FORCE/ACCELERATION CONTROL}

In order to control the translational force/acceleration of the multi-rotor, we need to convert the target acceleration $\ddot{\mathbf{X}}_d$ into the target attitude $\mathbf{q}_d$ and the target thrust $T_{t,d}$. Throughout this paper, notation $(*)_d$ denotes the desired value of the variable $*$. Also, we assume that the yaw $\psi$ of $\mathbf{q}$ always remains zero through a well-behaved independent controller to simplify the discussion. Now, we define $\mathbf{r}=[\theta\ \phi\ T_t]^T\in\mathbb{R}^{3\times1}$ as a set of states that needs to be controlled for generating the desired translational acceleration of the multi-rotor.

Once we choose $\mathbf{r} =[\theta\ \phi\ T_{t}]^T$ as a set of state variables to control the translational force/acceleration of multi-rotor, our next task should be finding a way to convert the desired acceleration $\ddot{\mathbf{X}}_d$ to $\mathbf{r}_d$. To figure out how to convert the signal, let us first investigate the relationship between $\mathbf{r}$ and $\ddot{\mathbf{X}}$.

\subsection{Relationship between $\mathbf{r}$ and $\tilde{\ddot{\mathbf{X}}}$}

In Equation (\ref{eq:dyn_full}), we have discussed the dynamics of the translational motion of multi-rotor. Going into detail, the corresponding translational dynamics are expressed as
\begin{equation}
\label{eq:dyn_trn_detail_rot}
m\ddot{\mathbf{X}}=-R(\psi)
\begin{bmatrix}
\cos{\phi}\sin{\theta}\\
-\sin{\phi}\\
\cos{\phi}\cos{\theta}
\end{bmatrix}
T_t+m\mathbf{g},
\end{equation}
where $R(\psi)\in\mathbb{R}^{3\times 3}$ is the yaw rotation matrix. Now, let us define a vector of state variables $\tilde{\ddot{\mathbf{X}}}$ named  the pseudo-acceleration vector  as
\begin{equation}
\label{eq:x_tilde_definition}
\tilde{\ddot{\mathbf{X}}}=
\begin{bmatrix}
\tilde{\ddot{x}}\\
\tilde{\ddot{y}}\\
\tilde{\ddot{z}}
\end{bmatrix}
=R^{-1}(\psi)
\left(\ddot{\mathbf{X}}-\mathbf{g}\right)=R^{-1}(\psi)\big({{1}\over{m}}\mathbf{F}\big).
\end{equation}
Applying Equation (\ref{eq:x_tilde_definition}) to (\ref{eq:dyn_trn_detail_rot}), we obtain the following relationship between $\mathbf{r}$ and $\tilde{\ddot{\mathbf{X}}}$:
\begin{equation}
\label{eq:accel-cmd}
m\tilde{\ddot{\mathbf{X}}}=-h(\phi,\theta)T_t=-
\begin{bmatrix}
\cos{\phi}\sin{\theta}\\
-\sin{\phi}\\
\cos{\phi}\cos{\theta}
\end{bmatrix}T_t.
\end{equation}

\subsection{Calculation of $\mathbf{r}_d$ from $\tilde{\ddot{\mathbf{X}}}_d$ considering dynamics}

From Equation (\ref{eq:accel-cmd}), we begin a discussion on how to calculate $\mathbf{r}_d$ based on $\tilde{\ddot{\mathbf{X}}}_d$. First, Equation (\ref{eq:accel-cmd}) yields the following expression on $\mathbf{r}$:
\begin{equation}
\label{eq:cmd-accel}
\mathbf{r}=
\begin{bmatrix}
\theta\\
\phi\\
T_t
\end{bmatrix}
=
\begin{bmatrix}
\arctan{\left({\tilde{\ddot{x}}}\over{\tilde{\ddot{z}}}\right)}\\
\arctan{\left(-{{\tilde{\ddot{y}}\cos{\theta}}\over{\tilde{\ddot{z}}}}\right)}\\
-{{m\tilde{\ddot{z}}}\over{\cos{\phi}\cos{\theta}}}
\end{bmatrix}.
\end{equation}
Equation (\ref{eq:cmd-accel}) represents  the required states $\mathbf{r}$ to generate such translational acceleration. From this, one might try to find the input to the controller to create the desired acceleration by replacing $\tilde{\ddot{\mathbf{X}}}$ and $\mathbf{r}$ with $\tilde{\ddot{\mathbf{X}}}_d$ and $\mathbf{r}_d$, respectively, as follows.
\begin{eqnarray}
\label{eq:rdes_candidate_theta}
\theta_d&\hspace{-.1cm}=\hspace{-.1cm}&\arctan{\left({\tilde{\ddot{x}}_d}\over{\tilde{\ddot{z}}_d}\right)}
\\
\label{eq:rdes_candidate_phi}
\phi_d&\hspace{-.1cm}=\hspace{-.1cm}&
\arctan{\left(-{{\tilde{\ddot{y}}_d\cos{\theta_d}}\over{\tilde{\ddot{z}}_d}}\right)}=\arctan{\left({{\tilde{\ddot{y}}_d}\over{\sqrt{\tilde{\ddot{x}}_d^2+\tilde{\ddot{z}}_d^2}}}\right)}
\\
\label{eq:rdes_candidate_T}
T_{t,d}&\hspace{-.1cm}=\hspace{-.1cm}&-{{m\tilde{\ddot{z}}_d}\over{\cos{\phi_d}\cos{\theta_d}}}=-m\sqrt{\tilde{\ddot{x}}_d^2+\tilde{\ddot{y}}_d^2+\tilde{\ddot{z}}_d^2}
\end{eqnarray}
However, this method can severely degrade control performance when multi-rotor is larger than a certain size as we discuss below.
\begin{figure}[t]
\begin{center}
\includegraphics[width=9cm]{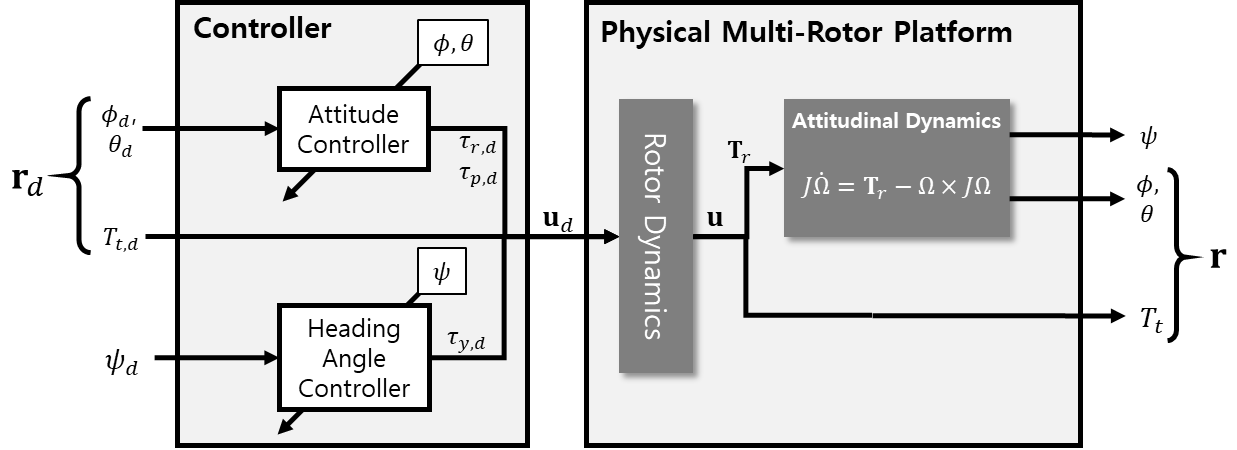}
\end{center}
\vspace{-0.5cm}
\caption{A block diagram of the relationship between $\mathbf{r}_d$ and $\mathbf{r}$, where $\mathbf{r}=[\theta, \phi, T_t]^T$, and $\mathbf{u}=[\tau_r, \tau_p, \tau_y, T_t]^T$.}
\label{gp:platform_structure}
\end{figure}

Fig. \ref{gp:platform_structure} shows the internal structure between $\mathbf{r}_d$ and $\mathbf{r}$. In this figure, we can see that $\phi_d$ and $\theta_d$ are realized to $\phi$ and $\theta$ through attitude controller, rotor dynamics, and attitude dynamics. In contrast, $T_{t,d}$ only passes through the rotor dynamics to become $T_t$. Here, we treat $\mathbf{u}_d=\mathbf{u}$, where $\mathbf{u}=[\mathbf{T}_r\ T_t]^T\in\mathbb{R}^{4\times1}$, since rotor dynamics are mostly negligible. Assuming that the attitude controller is properly designed, we can model the relationship between $\mathbf{r}_d$ and $\mathbf{r}$ as the following equation:
\begin{equation}
\label{eq:iorelationship}
\mathbf{r}(t)=
\begin{bmatrix}
\theta(t)\\
\phi(t)\\
T_t(t)
\end{bmatrix}
\approx
\begin{bmatrix}
\theta_d\left(t-\gamma_\theta\right)\\
\phi_d\left(t-\gamma_\phi\right)\\
T_{t,d}(t)
\end{bmatrix}.
\end{equation}
Here, $\gamma_*\in[0,\infty)$ are time-varying non-negative delay factors. Applying Equation (\ref{eq:iorelationship}) into (\ref{eq:accel-cmd}), we have
\begin{equation}
\label{eq:dynamics_false}
m\tilde{\ddot{\mathbf{X}}}=h\Big(\phi_d\big(t-\gamma_\phi\big),\theta_d\big(t-\gamma_\theta\big)\Big)T_{t,d}(t).\\
\end{equation}
In Equation (\ref{eq:dynamics_false}), the desired attitude and total thrust are realized asynchronously due to  $\gamma_\phi$ and $\gamma_\theta$. Applying Equation (\ref{eq:rdes_candidate_T}) to Equation (\ref{eq:dynamics_false}), the result is follows.
\begin{equation}
\label{eq:dynamics_false_detail}
\begin{bmatrix}
\tilde{\ddot{x}}(t)\\
\tilde{\ddot{y}}(t)\\
\tilde{\ddot{z}}(t)
\end{bmatrix}
=
\begin{bmatrix}
\left({{\cos{\phi_d(t-\gamma_\phi)}}\over{\cos{\phi_d(t)}\cos{\theta_d(t)}}}\tilde{\ddot{z}}_d(t)\right)\sin{\theta_d(t-\gamma_\theta)}\\
-\left({{1}\over{\cos{\phi_d(t)\cos{\theta_d(t)}}}}\tilde{\ddot{z}}_d(t)\right)\sin{\phi_d(t-\gamma_\phi)}\\
\left({{\cos{\phi_d(t-\gamma_\phi)\cos{\theta_d(t-\gamma_\theta)}}}\over{\cos{\phi_d(t)}\cos{\theta_d(t)}}}\right)\tilde{\ddot{z}}_d(t)
\end{bmatrix}
\end{equation}
In the $\tilde{\ddot{z}}(t)$ equation of Equation (\ref{eq:dynamics_false_detail}), the parenthesized part can continuously change if $\gamma_\phi$ and $\gamma_\theta$ are too large to be ignored. 
This indicates that $z$-directional control performance can be significantly reduced if the delay between the desired and actual attitude signals becomes large, for example in situations when the MOI of the multi-rotor increases, such as large multi-rotor or multi-rotor with large cargo.
{When the Z-directional control performance degrades, a high-level controller (e.g., position controller) or the operator may need to constantly modify the $\tilde{\ddot{z}}_d$ value to correct the poor Z-directional control performance.
As a result, this degrades the X and Y direction control performance because the values in parentheses of the $\tilde{\ddot{x}}(t)$ and $\tilde{\ddot{y}}(t)$ equations in \eqref{eq:dynamics_false_detail} also constantly change.}
The decline in control performance due to this control scheme will be shown in Fig. \ref{gp:pos_acc_comparison}.

To address this issue, we next consider two candidate solutions. 

\subsubsection{Solution candidate 1}
The first candidate is to time-synchronize the attitude and total thrust output by adding an artificial time delay to $T_{t,d}$ in Equation (\ref{eq:rdes_candidate_T}) as
\begin{equation}
\label{eq:rdes_candidate_T2}
T_{t,d}=-{{m\tilde{\ddot{z}}_d(t-\gamma_v)}\over{\cos{\phi_d(t-\gamma_\phi)}\cos\theta_d(t-\gamma_\theta)}}.
\end{equation}
Here, $\gamma_v$ is a delay element deliberately applied to $\tilde{\ddot{z}}_d$. 
Applying Equation (\ref{eq:rdes_candidate_T2}) to Equation (\ref{eq:dynamics_false}), the equation of motion is changed from Equation (\ref{eq:dynamics_false_detail}) to 
\begin{equation}
\label{eq:dynamics_false_detail2}
\begin{bmatrix}
\tilde{\ddot{x}}(t)\\
\tilde{\ddot{y}}(t)\\
\tilde{\ddot{z}}(t)
\end{bmatrix}
=
\begin{bmatrix}
\tan{\theta_d(t-\gamma_\theta)}\\
-{{\tan{\phi_d(t-\gamma_\phi)}}\over{\cos{\theta_d(t-\gamma_\theta)}}}\\
1
\end{bmatrix}
\tilde{\ddot{z}}_d(t-\gamma_v).
\end{equation}
Through Equations (\ref{eq:rdes_candidate_theta}), (\ref{eq:rdes_candidate_phi}) and (\ref{eq:iorelationship}), $\phi(t)$ and $\theta(t)$ can be described as
\begin{equation}
\label{eq:attitude-attitude_d relationship}
\left\{
\begin{array}{lr}
\theta(t)=\arctan{\left({{\tilde{\ddot{x}}_d(t-\gamma_\theta)}\over{\tilde{\ddot{z}}_d(t-\gamma_\theta)}}\right)}\\
\phi(t)=\arctan{\left(-{{\tilde{\ddot{y}}_d(t-\gamma_\phi)\cos{\theta_d(t-\gamma_\phi)}}\over{\tilde{\ddot{z}}_d(t-\gamma_\phi)}}\right)}.
\end{array}
\right.
\end{equation}
Let us assume that $\gamma_\phi$ and $\gamma_\theta$ have the same value of $\gamma_h$ since most multi-rotors have nearly the same roll and pitch behavior due to the symmetrical mechanical structure. 
Then, Equation (\ref{eq:dynamics_false_detail2}) with Equations (\ref{eq:iorelationship}) and (\ref{eq:attitude-attitude_d relationship}) becomes as
\begin{equation}
\label{eq:tdynamics_final_candidate}
\begin{bmatrix}
\tilde{\ddot{x}}(t)\\
\tilde{\ddot{y}}(t)\\
\tilde{\ddot{z}}(t)
\end{bmatrix}
=
\begin{bmatrix}
\tilde{\ddot{x}}_d(t-\gamma_h)\big({{{\tilde{\ddot{z}}_d(t-\gamma_v)}\over{\ddot{z}}_d(t-\gamma_h)}}\big)\\
\tilde{\ddot{y}}_d(t-\gamma_h)\big({{{\tilde{\ddot{z}}_d(t-\gamma_v)}\over{\ddot{z}}_d(t-\gamma_h)}}\big)\\
\tilde{\ddot{z}}_d(t-\gamma_v)
\end{bmatrix}.
\end{equation}
Now, we can solve the problem in Equation (\ref{eq:dynamics_false_detail}) by setting $\gamma_v$ equal to $\gamma_h$.
However, this method is not easily applicable in a real-world situation because it is difficult to determine the value of $\gamma_h$ that changes continuously during the flight.
Therefore, the control method through Equation (\ref{eq:rdes_candidate_T2}) cannot be a practical method.

\subsubsection{Solution candidate 2}
Alternatively, we can find a reasonable solution that is applicable in the real world by selectively delaying $\phi_d(t)$ and $\theta_d(t)$ in Equation (\ref{eq:rdes_candidate_T2}) by $\gamma_\phi$ and $\gamma_\theta$, but keeping $\gamma_v$ at zero.
As we can see from Equation (\ref{eq:iorelationship}), the values of $\phi_d(t)$ and $\theta_d(t)$ delayed by $\gamma_\phi$ and $\gamma_\theta$ seconds are $\phi(t)$ and $\theta(t)$. 
Applying this idea to Equation (\ref{eq:rdes_candidate_T2}), we can obtain $T_{t,d}$ as
\begin{equation}
\label{eq:Tfd_determination}
T_{t,d}=-{{m\tilde{\ddot{z}}_d(t)}\over{\cos{\phi(t)}\cos{\theta(t)}}},
\end{equation}
where the values $\phi(t)$ and $\theta(t)$ can be measured from the built-in inertial measurement unit (IMU) sensor. 
Then, by setting $\gamma_v$ to zero, we can determine the input/output relationship of the translational accelerations dynamics of the multi-rotor as
\begin{equation}
\label{eq:tdynamics_final}
\tilde{\ddot{\mathbf{X}}}=
\begin{bmatrix}
\tilde{\ddot{x}}_d(t-\gamma_h)\left({{\tilde{\ddot{z}}_d(t)}\over{\tilde{\ddot{z}}_d(t-\gamma_h)}}\right)\\
\tilde{\ddot{y}}_d(t-\gamma_h)\left({{\tilde{\ddot{z}}_d(t)}\over{\tilde{\ddot{z}}_d(t-\gamma_h)}}\right)\\
\tilde{\ddot{z}}_d(t)
\end{bmatrix}\approx
\begin{bmatrix}
\tilde{\ddot{x}}_d(t-\gamma_h)\\
\tilde{\ddot{y}}_d(t-\gamma_h)\\
\tilde{\ddot{z}}_d(t)
\end{bmatrix}\;,
\end{equation}
where we assume ${{\tilde{\ddot{z}}_d(t)}\over{\tilde{\ddot{z}}_d(t-\gamma_h)}}\approx1$.
This assumption is valid in most cases, except in situations where the change in target vertical acceleration is abnormally {large and} rapid.

\begin{figure*}[t]
\begin{center}
\includegraphics[width=18cm]{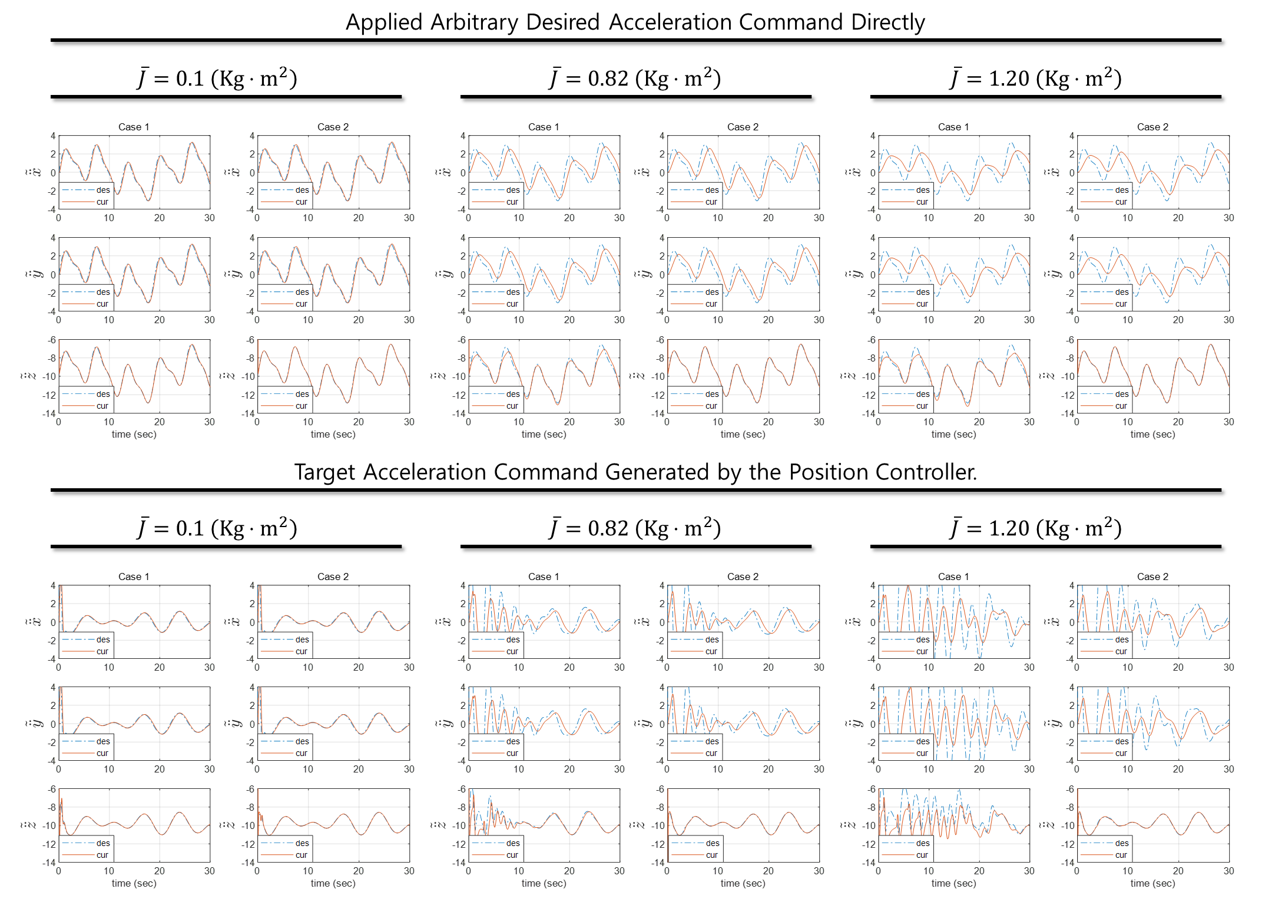}
\end{center}
\vspace{-0.5cm}
\caption{{[Simulation] A comparison of cases where acceleration command is converted into a target attitude and a thrust signal using Equations (\ref{eq:rdes_candidate_theta}), (\ref{eq:rdes_candidate_phi}) and (\ref{eq:rdes_candidate_T}) (Case 1), and using Equations (\ref{eq:rdes_candidate_theta}), (\ref{eq:rdes_candidate_phi}) and (\ref{eq:Tfd_determination}) (Case 2) for multi-rotors with different MOI. Acceleration motions are simulated for two scenarios : in the first scenario, an arbitrary target acceleration command is applied (top), and the target acceleration is generated via a position controller that tracks the predefined desired trajectory in the second scenario (bottom).}}
\label{gp:pos_acc_comparison}
\end{figure*}

Through the control techniques of solution candidate 2 \big(Equations (\ref{eq:rdes_candidate_theta}), (\ref{eq:rdes_candidate_phi}) and (\ref{eq:Tfd_determination})\big), we obtained a three-dimensional translational acceleration control method applicable to actual multi-rotor control.
In order to compare the performance of multi-rotor control using Equations (\ref{eq:rdes_candidate_T}) and (\ref{eq:Tfd_determination}), a brief simulation is conducted as shown in Fig. \ref{gp:pos_acc_comparison}. 

{The simulation shows the comparison of the target acceleration tracking performance of Case 1 with Equations (\ref{eq:rdes_candidate_theta}), (\ref{eq:rdes_candidate_phi}), (\ref{eq:rdes_candidate_T}) and Case 2 with Equations (\ref{eq:rdes_candidate_theta}), (\ref{eq:rdes_candidate_phi}), (\ref{eq:Tfd_determination}).
The upper set of figures show the acceleration tracking performance of Cases 1 and 2 with arbitrary acceleration command.
Here, we can see that there are no differences in performance between Cases 1 and 2 when MOI of the multi-rotor has small value of 0.1.
On the other hand, when the MOI of the multi-rotor increases, both Cases 1 and 2 show delayed responses in the X and Y direction acceleration tracking as expected.
However, we can observe that the Z-directional performance of the Case 2 remains the same regardless of the magnitude of the MOI, unlike Case 1 where the performance degradation is observed.
The effect of the decline in Z-directional control performance on the system is evident when controlling the position of the multi-rotor.
The bottom set of figures is the situation where the high-level  position controller generates the desired acceleration command to track the predefined trajectory.
In Case 1, we can observe a decrease in acceleration tracking performance in both the X and Y directions as well as the Z direction as the MOI increases.
On the other hand, in Case 2, the Z-directional control performance remains constant regardless of the MOI of the platform, stabilizing the X and Y-directional control performance faster than Case 1.}

This phenomenon can be understood in other ways by considering the role of the denominator term of the $T_t$ equation in Equation (\ref{eq:cmd-accel}), which is to compensate for the reduction of the vertical thrust component in the sense of inertial coordinates when the multi-rotor is tilted.
When $T_{t,d}$ is calculated based on the desired attitude as Equation (\ref{eq:rdes_candidate_T}), the situation is similar to compensating for the future event after $\gamma_h$ seconds. 
Instead, it is intuitive to use the current attitude as in Equation (\ref{eq:Tfd_determination}) to correct the vertical thrust reduction. 
From the flight results using Equation (\ref{eq:Tfd_determination}) in Fig. \ref{gp:pos_acc_comparison}, we can confirm that the control performance in all directions is satisfactory.

\section{DISTURBANCE OBSERVER}

\begin{figure*}[t]
\begin{center}
\includegraphics[width=18cm]{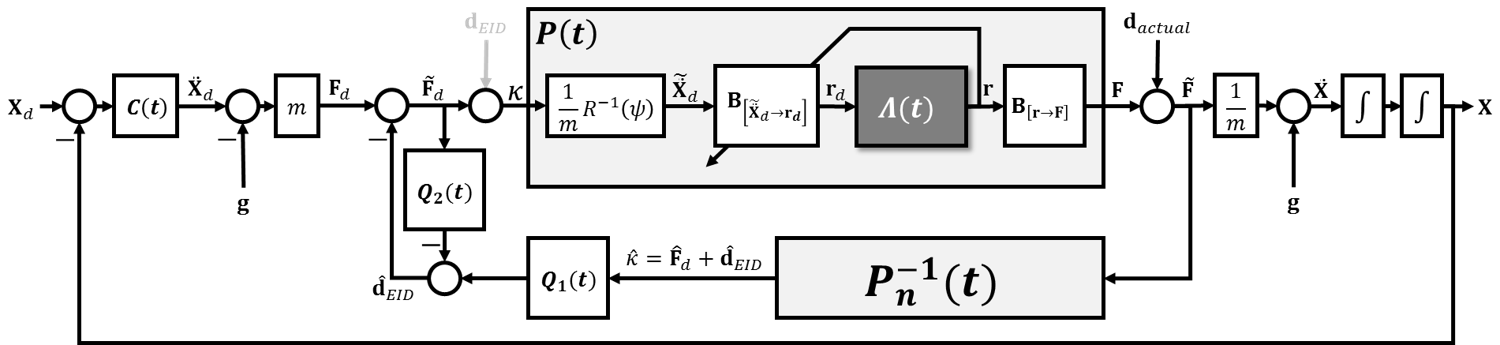}
\end{center}
\vspace{-0.5cm}
\caption{Overall system diagram with DOB structure. $C(t)$: Outer-loop controller, $\mathbf{F}_d$: Desired translational force vector, $\tilde{\mathbf{F}}_d$: Sum of $\mathbf{F}_d$ and disturbance cancellation signal $-\hat{\mathbf{d}}_{EID}$, $\mathbf{B}\scriptscriptstyle{[\tilde{\ddot{\mathbf{X}}}_d\rightarrow\mathbf{r}_d]}$: $\tilde{\ddot{\mathbf{X}}}_d$ to $\mathbf{r}_d$ translator \big(eq. (\ref{eq:rdes_candidate_theta}), (\ref{eq:rdes_candidate_phi}), (\ref{eq:Tfd_determination})\big), $\it{\Lambda(t)}$: Plant dynamics \big(Fig. \ref{gp:platform_structure}, Eq. (\ref{eq:dyn_full})\big), $\mathbf{B}\scriptscriptstyle{[\mathbf{r}\rightarrow\mathbf{F}]}$: $\mathbf{r}$ to $\mathbf{F}$ translator \big(Eq. (\ref{eq:x_tilde_definition}), (\ref{eq:accel-cmd})\big), $\mathbf{F}$: Force vector generated by the multi-rotor, $\tilde{\mathbf{F}}$: Sum of $\mathbf{F}$ and actual disturbance $\mathbf{d}_{actual}$, ${P_n(t)}$: Nominal model of ${P(t)}$, ${Q_{1,2}(t)}$: $Q$-filters for DOB. }
\label{gp:full_diagram}
\end{figure*}

\begin{figure}[t]
\begin{center}
\includegraphics[width=8cm]{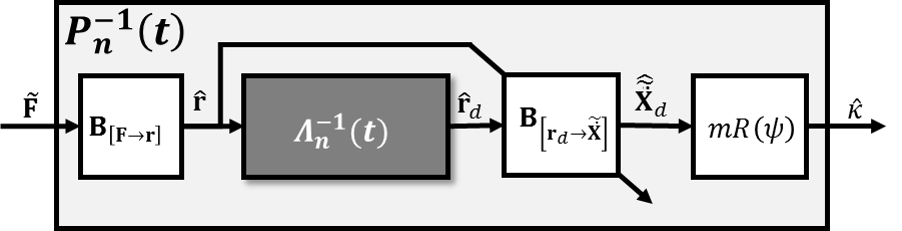}
\end{center}
\vspace{-0.5cm}
\caption{Configuration of ${P_n^{-1}(t)}$. The ${P_n^{-1}(t)}$ block is composed of the opposite order of ${P(t)}$, where ${\it{\Lambda}}_n(t)$ is the nominal model of ${\it{\Lambda}}(t)$.}
\label{gp:FtoK}
\end{figure}

 External disturbances applied to multi-rotor act not only in the form of translational disturbances but also in the form of rotational torques. However, given that a number of solutions for overcoming the rotational torque disturbances \cite{Attitude_DOB_1}$\sim$\cite{suseong_DOB} have already been proposed, this section concerns only translational disturbances applied to the system for straightforward discussion and analysis.

\subsection{An overview of the disturbance-merged overall system}

Fig. \ref{gp:full_diagram} shows the overall configuration of the system. First, the position controller ${C(t)}$ generates the target acceleration input $\ddot{\mathbf{X}}_d$. This signal is then transformed into the target force input $\mathbf{F}_d$ through the following force-acceleration relationship:
\begin{equation}
\label{eq:X-F relationship}
\mathbf{F}=m(\ddot{\mathbf{X}}-\mathbf{g}).
\end{equation}
Then, $\mathbf{F}_d$ signal passes through ${{1}\over{m}}R^{-1}(\psi)$ block to transform the signal into the $\tilde{\ddot{\mathbf{X}}}_d$ \big(refer Equation (\ref{eq:x_tilde_definition})\big). The signal $\tilde{\ddot{\mathbf{X}}}_d$ then passes through the $\mathbf{B}\scriptscriptstyle{[\tilde{\ddot{\mathbf{X}}}_d\rightarrow\mathbf{r}_d]}$ block, which converts the target acceleration $\tilde{\ddot{\mathbf{X}}}_d$ to $\mathbf{r}_d$, the  input to the multi-rotor controller,  based on Equations (\ref{eq:rdes_candidate_theta}), (\ref{eq:rdes_candidate_phi}) and (\ref{eq:Tfd_determination}). 
Once $\mathbf{r}_d$ passes through the dynamics described in Fig. \ref{gp:platform_structure} and outputs $\mathbf{r}$, it passes through $\mathbf{B}\scriptscriptstyle{[\mathbf{r}\rightarrow\mathbf{F}]}$ block to produce $\mathbf{F}$ \big(refer Equation (\ref{eq:x_tilde_definition}) and (\ref{eq:accel-cmd})\big). 
Right after $\mathbf{F}$ is generated, the external disturbance force $\mathbf{d}_{actual}$ immediately compromises the thrust and results in $\tilde{\mathbf{F}}$ and $\ddot{\mathbf{X}}$.

\subsection{Disturbance observer}

In Fig. \ref{gp:full_diagram}, the translational force disturbance $\mathbf{d}_{actual}$ is combined with $\mathbf{F}$ to become $\tilde{\mathbf{F}}$. However, canceling $\mathbf{d}_{actual}$ is only possible by adding an appropriate disturbance cancellation term to the $\mathbf{F}_d$ signal. Therefore, it is preferable to assume that there is an equivalent input disturbance $\mathbf{d}_{EID}$ that has the same effect on the system as $\mathbf{d}_{actual}$ \cite{EID}. Then $\mathbf{d}_{actual}$ is replaced by $\mathbf{d}_{EID}$, making  $\mathbf{F}=\tilde{\mathbf{F}}$. As we can see in Fig. \ref{gp:full_diagram}, the $\mathbf{d}_{EID}$ signal is merged into $\tilde{\mathbf{F}}_d$, which is the translational acceleration control input with disturbance cancellation signal. Now, let us construct the DOB based on the above settings.

\subsubsection{$\mathbf{d}_{EID}$ estimation algorithm}

For the estimation of $\mathbf{d}_{EID}$,  we first estimate $\kappa$ the sum of $\tilde{\mathbf{F}}_d$ and $\mathbf{d}_{EID}$ by
\begin{equation}
\label{eq:k_hat}
\hat{\kappa}(s)=\hat{\tilde{\mathbf{F}}}_d(s)+\hat{\mathbf{d}}_{EID}(s)={P_n^{-1}}(s)\tilde{\mathbf{F}}(s).
\end{equation}
We can easily achieve the $\tilde{\mathbf{F}}$ signal from Equation (\ref{eq:X-F relationship}) where $\ddot{\mathbf{X}}$ is measured by the IMU sensor. The transfer function ${P_n(s)}$ is the nominal model of ${P(s)}$, and $\hat{(*)}$ is the representation of the estimation of $(*)$ signal throughout this paper. Once we estimate $\hat{\kappa}$, we then obtain $\hat{\mathbf{d}}_{EID}$ by 
\begin{equation}
\label{eq:dhat_eid}
\hat{\mathbf{d}}_{EID}={Q_1}(s)\hat{\kappa}(s)-{Q_2}(s)\tilde{\mathbf{F}}_d(s).
\end{equation}
The signal $\hat{\kappa}(s)$ passes through the ${Q_1}$ block, which is basically a low pass filter, to overcome both the causality violation issue due to the improperness of ${P_n^{-1}}(s)$ and the potential instability issue caused by the non-minimum phase characteristic of ${P_n}(s)$. The filter ${Q_2}(s)$ is used to match the phase with ${Q_1}(s)\hat{\kappa}(s)$ signal. In the end, we generate a disturbance-compensating control input $\tilde{\mathbf{F}}_d$ by 
\begin{equation}
\label{eq:Ftilde_d}
\tilde{\mathbf{F}}_d=\mathbf{F}_d-\hat{\mathbf{d}}_{EID}.
\end{equation}
This makes $\kappa$ become
\begin{equation}
\label{eq:kappa}
\kappa=\tilde{\mathbf{F}}_d+\mathbf{d}_{EID}=\mathbf{F}_d-\hat{\mathbf{d}}_{EID}+\mathbf{d}_{EID}\approx\mathbf{F}_d.
\end{equation}
The most important factor in the $\mathbf{d}_{EID}$ estimation process is the proper design of ${P_n}$ and ${Q}$. Of these, ${Q}$ is deeply related to the stability of the system and will be discussed in more detail in the next section. In the remainder of this section, we first discuss the design of the nominal model ${P_n}$ and then explain the structure of the $Q$-filter.

\subsubsection{Nominal model ${P_n}$}

The internal structure of ${P_n^{-1}}(t)$ is described as in Fig. \ref{gp:FtoK}, all of which are simple conversion blocks except for the ${{\it{\Lambda}}}^{-1}_n{\it(t)}$ block. The block ${\it{\Lambda}}(t)$ is the relationship between $\mathbf{r}_d$ and $\mathbf{r}$ depicted in Fig. \ref{gp:platform_structure}. The $\it{\Lambda_n(s)}$ is constructed from two parts: attitude and thrust dynamics. We denote these as $\it{\Lambda_{n,a}}(s)$ and $\it{\Lambda_{n,t}}(s)$ respectively. 

As we see from Fig. \ref{gp:platform_structure}, $\it{\Lambda}_{n,a}(s)$ is constructed with attitude controller, rotor dynamics and attitudinal dynamics. Since rotor dynamics can be ignored, we only need to find the transfer function of the attitudinal dynamics and attitude controller. For attitude dynamics, let us refer to Equation (\ref{eq:dyn_rot_simplify}) and express it as
\begin{equation}
\label{eq:atti.-dyn_freq.}
{{\mathbf{q}_i(s)}\over{\mathbf{T}_{r,i}(s)}}={{1}\over{J_is^2}},
\end{equation}
where $i=1,2,3$ represent $\phi$, $\theta$, $\psi$ axis, respectively. For attitude control, PD control in the following form is used.
\begin{equation}
\label{eq:PID}
{{\mathbf{T}_{r,i}(s)}\over{\mathbf{q}_{i,d}(s)-\mathbf{q}_i(s)}}={P_i+D_is}
\end{equation}
The parameters $P_i$, $D_i$ represent control gains in each attitude component. Then, the overall transfer function ${\it{\Lambda}}_{n,a,i}$ between desired and current attitude becomes
\begin{equation}
\label{eq:lambda_na}
{\it{\Lambda}}_{n,a,i}(s)={{\mathbf{q}_i(s)}\over{\mathbf{q}_{i,d}(s)}}={{D_is+P_i}\over{J_is^2+D_is+P_i}}.
\end{equation}

In the case of ${\it{\Lambda}}_{n,t}$, the only dynamics involved is rotor dynamics, which we decided to neglect. Thus, it can be expressed as 
\begin{equation}
\label{eq:lambda_nt}
{\it{\Lambda}}_{n,t}(s)=1. 
\end{equation}
Now, we can construct the transfer matrix for ${\it{\Lambda}}_n=diag({\it{\Lambda}}_{n,1},{\it{\Lambda}}_{n,2},{\it{\Lambda}}_{n,3})$ using Equations (\ref{eq:lambda_na}) and (\ref{eq:lambda_nt}) as
\begin{equation}
\label{eq:lambda_n}
{\it{\Lambda}}_n(s)=
\begin{bmatrix}
{\it{\Lambda}}_{n,a,2}(s) & 0 & 0\\
0 & {\it{\Lambda}}_{n,a,1}(s) & 0\\
0 & 0 & {\it{\Lambda}}_{n,t}(s)
\end{bmatrix}.
\end{equation}

Equation (\ref{eq:lambda_n}) is a detailed representation of the relationship between $\mathbf{r}_d$ and $\mathbf{r}$, which was introduced in Equation (\ref{eq:iorelationship}). 
On the other hand, $P_n$, which defines the nominal relationship between $\kappa$ and $\mathbf{F}$ \big(or $\tilde{\ddot{\mathbf{X}}}_d$ and $\tilde{\ddot{\mathbf{X}}}$\big), was introduced in Equation (\ref{eq:tdynamics_final}) \big(refer Equation (\ref{eq:x_tilde_definition}) for the relationship between $\tilde{\ddot{\mathbf{X}}}$ and $\mathbf{F}$\big). Here, we can see that both Equations (\ref{eq:iorelationship}) and (\ref{eq:tdynamics_final}) have the same input/output characteristics with time delay of $\gamma_h$ for the first and second channels and no time delay for the third channel. 
Therefore, we can conclude that ${\it{\Lambda}}_n(s)$ in Equation (\ref{eq:lambda_n}) is also the transfer function between $\kappa$ and $\mathbf{F}$ as well as between $\mathbf{r}_d$ and $\mathbf{r}$, which is 
\begin{equation}
\label{eq:PneqLn}
{P}_n(s)={\it{\Lambda}}_n(s).
\end{equation}

\subsubsection{$Q$-filter design}
In $Q$-filter design, we choose to make ${Q_1}(s){\it{\Lambda}}_n^{-1}(s)$, which is now identical to ${Q_1}(s){P}_n^{-1}(s)$, a proper function with relative degree of 1. Since ${P}_n(s)$ is composed of three channels in $X$, $Y$ and $Z$ directions, we need to design three separate $Q$-filters. As shown in Equation (\ref{eq:lambda_na}), ${\it{\Lambda}}_{n,1}(s)$\big($={\it{\Lambda}}_{n,a,2}(s)$\big) and ${\it{\Lambda}}_{n,2}(s)$\big($={\it{\Lambda}}_{n,a,1}(s)$\big) among the three transfer functions of ${\it{\Lambda}}_n(s)$ are systems with a relative degree of 1. The thrust transfer function ${\it{\Lambda}}_{n,3}(s)$\big($={\it{\Lambda}}_{n,t}(s)$\big) has a relative degree of 0, as can be seen from Equation (\ref{eq:lambda_nt}). Therefore, the $Q$-filters for making ${Q_1}(s){\it{\Lambda}}_n^{-1}(s)$ with a relative degree of 1 are designed as
\begin{equation}
\label{eq:Q_1}
{Q_1}(s)=diag\big({Q_{1,h}}(s),{Q_{1,h}}(s),{Q_{1,v}}(s)\big),
\end{equation}
\begin{equation}
\label{eq:Q-filter1}
{Q_{1,h}}(s)={{1}\over{(\tau_1 s)^2+\zeta(\tau_1 s)+1}},
\end{equation}
\begin{equation}
\label{eq:Q_filter2}
{Q_{1,v}}(s)={{1}\over{(\tau_2 s)+1}},
\end{equation}
where ${Q_{1,h}}$ and ${Q_{1,v}}$ are $Q$-filters corresponding to the horizontal ($\it{\Lambda}_{n,a}$) and vertical ($\it{\Lambda}_t$) models respectively. The symbol $\tau$ is the time constant and $\zeta$ is the damping ratio of the filter. The filter ${Q_1}$ is designed to have a gain of 1 when $s=0$ \cite{DOB_Q}. The filter ${Q_2}$ is set to ${Q_2}={Q_1}$, to easily achieve the purpose of phase matching.

\section{STABILITY ANALYSIS}

The design of $Q$-filter in the  DOB structure should be based on rigorous stability analysis to ensure the overall stability. In particular, we note that there is always a difference between the nominal model $P_n(s)$ and the actual model $P(s)$, due to various uncertainties and applied assumptions. 

Although the small-gain theorem (SGT) \cite{DOB_sjlazza} can still be a tool for stability analysis, the SGT analysis based on the largest singular value among uncertainties is  likely to yield overly conservative results especially if multiple uncertain elements are involved. Instead, we use structured singular value analysis, or $\mu$-analysis \cite{SGTnmu, sariyildiz2013analysis, kim2018robust}, to reflect the combined effects of uncertainties. 

\subsection{Modeling of ${P}(s)$ considering uncertainties}
The multi-rotor's actual transfer function ${P}(s)$ between $\kappa$ and $\mathbf{F}$ in Fig. \ref{gp:full_diagram} is
\begin{equation}
\label{eq:P}
{P}(s)=diag\big({P}_{1}(s),{P}_{2}(s),{P}_3(s)\big).
\end{equation}
Here, $P_1$, $P_2$ and $P_3$ represent the input/output translational force relationship in the $X$, $Y$, and $Z$ directions, respectively. 
This research considers a small but nonzero DC-gain error, parametric error and phase shift error between ${P}_n(s)$ and ${P}(s)$. 
Then each ${P}_{j}(s)$ can be expressed as the following equation:
\begin{equation}
\label{eq:P_model}
\begin{split}
{P}_{j}(s)&=K_{j}{P}_{n,j}(s)e^{-\delta_{j} s}\\
&=K_{j}{\it{\Lambda}}_{n,j}(s)e^{-\delta_{j} s}={P}_{p,j}(s)\Gamma_{j}(s),
\end{split}
\end{equation}
where $j=1,2,3$ represent $X$, $Y$, $Z$ axis. The symbols $K_{j},\delta_j\in\mathbb{R}$ represent the uncertain variable gain and time delay parameters, respectively. The nominal transfer function ${P}_{n,j}$ can be replaced by $\it{\Lambda}_{n,j}$ based on Equation (\ref{eq:PneqLn}). The portion containing only the parametric uncertainty is denoted by ${P}_{p,j}(s)=K_{j}{\it{\Lambda}}_{n,j}(s)$, and the time delay uncertainty is denoted by $\Gamma_{j}(s)=e^{-\delta_{j} s}$. 

In Equation (\ref{eq:P_model}), each ${P}_{j}(s)$ contains three uncertain variables, which are $K_{j}$, $J_j$ and $\delta_{j}$. In the case of $K_j$, we define $K_j$ as 
\begin{equation}
\label{eq:K_j}
K_j=1+K_{\Delta,j},
\end{equation}
where $K_{\Delta,j}\in\mathbb{R}$ is the error value of $K_j$. In the case of $J_j$, determining the actual value of $J_j$ is difficult compared to other physical quantities. We also define $J_j$ in the same manner as $K_j$ for the convenience of analysis as 
\begin{equation}
\label{eq:J_j}
J_j=\bar{J}_j(1+J_{\Delta,j}),
\end{equation}
where $\bar{J}_j, J_{\Delta,j}\in\mathbb{R}$ are the nominal and error values of $J_j$. Because the term $\Gamma_{j}(s)$ containing $\delta_j$ is of an irrational form that is not suitable for analysis, we use an analytic approximation of the uncertain time-delay $\Gamma_{j}(s)$ to a rational function with unmodeled dynamic uncertainty \cite{kim2018robust}. First, we change the representation of the $P_j(s)$ model to a multiplicative uncertainty form that combines parametric uncertainties and unmodeled time-delay uncertainty as follows:
\begin{equation}
\label{eq:unmodeled_plant}
\begin{split}
{P}_{j}(s)={P}_{p,j}(s)\big(1+\Delta_{\delta,j}(s)W_{\delta,j}(s)\big),\\
||\Delta_{\delta,j}(s)||_{\infty} \leq 1.
\end{split}
\end{equation}
\begin{figure}[t]
\begin{center}
\includegraphics[width=8cm]{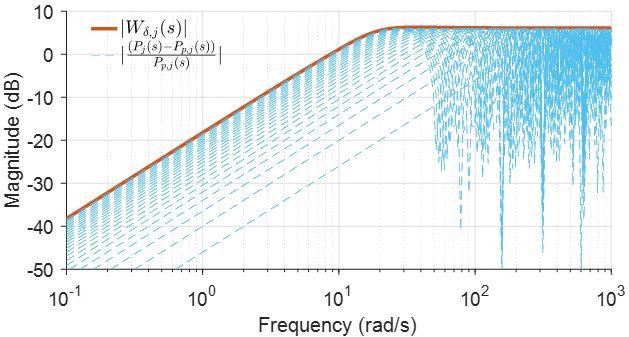}
\end{center}
\vspace{-0.5cm}
\caption{Bode magnitude plots of $\Gamma_j(s)-1$ expressed by varying $\delta_j$ from $-0.12$ to $0.12$ (blue dashed line), maximum uncertainty $W_{\delta,j}(s)$ (red solid line).}
\label{gp:W(s)}
\end{figure}
A complex unstructured uncertainty $\Delta_{\delta,j}\in\mathbb{C}$ corresponds to unknown time delay $\delta_j$, and $W_{\delta,j}(s)$ is the maximum uncertainty that can be caused by $\Gamma_{j}(s)$. Here, we can obtain $W_{\delta,j}(s)$ using Equation (\ref{eq:P_model}) as
\begin{equation}
\label{eq:W(s)}
W_{\delta,j}(s)=\underset{\delta_j}\max{\Big|{{{P}_{j}(s)-{P}_{p,j}(s)}\over{{P}_{p,j}(s)}}\Big|}=\underset{\delta_j}\max|\Gamma_{j}(s)-1|.
\end{equation}
The maximum value of $|\Gamma_{j}(j\omega)-1|$ for each $\omega$ can be found using Euler's formula as
\begin{equation}
\label{eq:W_euler}
\underset{\delta_j}\max|\Gamma_{j}(j\omega)-1|=
\underset{\delta_j}\max\sqrt{\big(\cos{(\omega\delta_{j})-1\big)^2+\big(\sin{(\omega\delta_{j})\big)^2}}}
\end{equation}
where
\begin{equation}
\label{eq:cossin}
\Gamma_{j}(j\omega)=e^{\delta_{j}(j\omega)}=\cos{(\omega\delta_{j})}+j\sin{(\omega\delta_{j})}.
\end{equation}
As a result of analyzing a large amount of actual experimental data, we confirmed that the time delay between ${P}_n(s)$\big($={\it{\Lambda}}_n(s)$\big) and ${P}(s)$ does not exceed 0.1 second in all three channels. We put 20 percent margin  so that $|\delta_j|\leq 0.12$. Fig. \ref{gp:W(s)} is multiple Bode magnitude plots of $|\Gamma(s)-1|$ generated by varying $\delta$ from $-0.12$ to $+0.12$. From Fig. \ref{gp:W(s)}, we can extract
\begin{equation}
\label{eq:W_max(S)}
W_{\delta,j}(s)={{2.015s^3+52.88s^2+431.6s+0.415}\over{s^3+36.7s^2+606.8s+3521}}
\end{equation}
for all $j$, which is the upper boundary of $|\Gamma_{j}(s)-1|$ sets represented by the red solid line.

The uncertainties of $K_{j}$ and $J_j$ can also be modeled in the same manner as in Equation (\ref{eq:unmodeled_plant}) as
\begin{equation}
\label{eq:Pp_rewritten}
\left\{
\begin{array}{lr}
P_{p,j}(s)=K_j {\it{\Lambda}}_{n,j}(s)={\it{\Lambda}}_{n,j}(1+{\Delta}_{K,j}{W}_{K,j})\\
{\it{\Lambda}}_{n,j}(s)={\it{\Lambda}}_{n,n,j}(s)(1+{\Delta}_{J,j}{W}_{J,j}),
\end{array}
\right.
\end{equation}
where $||{\Delta}_{K,j}||_\infty,||{\Delta}_{J,j}||_\infty\leq1$. The transfer function ${\it{\Lambda}}_{n,n,j}$ is basically the same as ${\it{\Lambda}}_{n,j}$, except that $J$ in Equation (\ref{eq:lambda_na}) is replaced to the nominal MOI value $\bar{J}$. The transfer functions $W_{K,j}$ and $W_{J,j}$ are
\begin{eqnarray} 
\label{eq:Wk,Wj}
W_{K,j}&\hspace{-.1cm}=\hspace{-.1cm}&\max|K_{\Delta,j}|\\
W_{J,j}&\hspace{-.1cm} =\hspace{-.1cm}& \left\{
\begin{array}{ll} \underset{J_{\Delta,j}}\max \Big|{{-\bar{J}_jJ_{\Delta,j}s^3}\over{\bar{J}_j(1+J_{\Delta,j})s^3+D_js^2+P_js+I_j}}\Big| &
 (j=1,2)\\
0 & (j=3).
\end{array} \nonumber
\right.
\end{eqnarray}

\subsection{$\tau$-determination through $\mu$-analysis}

\subsubsection{$\mu$-robust stability analysis}
In \cite{muDoyle}, the structured singular value $\mu$ is defined as
\begin{equation}
\label{eq:mu}
\mu_{\Delta}(M_{11})={{1}\over{\underset{\Delta\in\mathbf{\Delta}}\min\big(\bar{\sigma}(\Delta):\det(I-M_{11}\Delta)=0\big)}}
\end{equation}
where $\Delta$ is a complex structured block-diagonal unmodeled uncertainty block which gathers all model uncertainties \cite{mu-analysis}. Following the common notation, the symbol $\mathbf{\Delta}$ represents a set of all stable transfer matrices with the same structure (full, block-diagonal, or scalar blocks) and  nature (real or complex) as $\Delta$. The $\bar{\sigma}\big(\Delta\big)$ is the maximum singular value of uncertainty block $\Delta$. The matrices $M$ and $\Delta$ are defined by collapsing the simplified overall system to upper LFT uncertainty description as 
\begin{equation}
\label{eq:LFT uncertainty description}
\begin{bmatrix}
z\\
y
\end{bmatrix}
=
\begin{bmatrix}
M_{11} & M_{12}\\
M_{21} & M_{22}
\end{bmatrix}
\begin{bmatrix}
w\\
r
\end{bmatrix}
,\ w=\Delta z,
\end{equation}
where $M$ is the known part of the system, $r$ is a reference input and $y$ is an output of the overall system. In the theory of the $\mu$-analysis, it is well-known that the system is robustly stable if $\mu$ satisfies the following conditions
\begin{equation}
\label{eq:mu_stable_condition}
\mu_\Delta(M_{11})<1,\ \forall\omega
\end{equation}
\cite{SGTnmu}\cite{muDoyle}. 

\begin{figure}[t]
\begin{center}
\includegraphics[width=8.5cm]{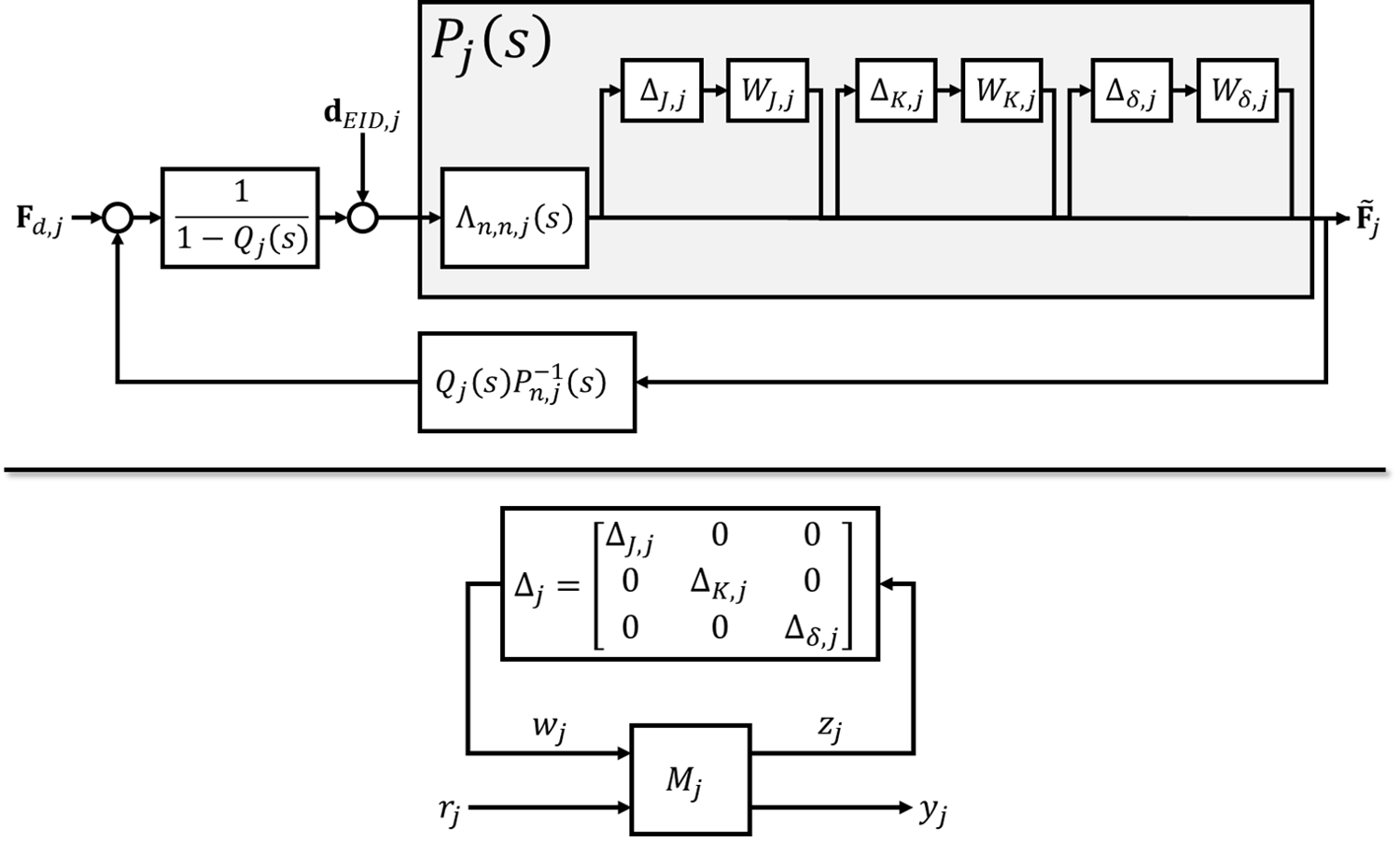}
\end{center}
\vspace{-0.5cm}
\caption{Compressed block digram of the DOB-included transfer function from $\mathbf{F}_{d,j}$ to $\tilde{\mathbf{F}}_j$, whose original form was shown in Fig. \ref{gp:full_diagram} (top), further collapsed form expressed as a nominal closed-loop system $M_j$ and a complex unstructured uncertainty block $\Delta_j$ as in Equation (\ref{eq:Delta_ourcase}) (bottom).}
\label{gp:lft_model}
\end{figure}

The $\mu$-analysis is performed separately for each channel of $X$, $Y$, $Z$ thanks to the structure of the platform described by Equation (28), but since $X$ and $Y$ channels are composed of the same structure, they share the identical analysis result. As we can see from Fig. \ref{gp:lft_model}, the system is collapsed in the form of Equation (\ref{eq:LFT uncertainty description}) by using MATLAB's Robust Control Toolbox$^{\textrm{TM}}$, where $r_j=[\mathbf{F}_{d,j}\ \mathbf{d}_{EID,j}]^T\in\mathbb{R}^{2\times1}$ and $y_j=\tilde{\mathbf{F}}_j\in\mathbb{R}$ in our case. As a reminder, subscript $j$ refers to each channel of $X$, $Y$, and $Z$. Also, structured uncertainty ${\Delta}_j(s)\in\mathbb{C}^{3\times3}$ is constructed as
\begin{equation}
\label{eq:Delta_ourcase}
{\Delta}_j(s)=diag\big({\Delta}_{J,j}(s), {\Delta}_{K,j}(s), {\Delta}_{\delta,j}(s)\big),
\end{equation}
which includes unmodeled MOI uncertainty, time and gain uncertainty in our system.
\subsubsection{Results of analysis}

\begin{table}[b]
\renewcommand{\arraystretch}{1.3}
\caption{PHYSICAL QUANTITIES AND CONTROLLER GAINS.}
\label{table_1}
\centering
\begin{tabular}{ |l|l|l|l| }
\hline
Name							&Value				& Name		    		& Value\\ \hline\hline
$P_{\phi,\theta}$ 				& 3 				& Mass				    & 3.24 \si{Kg}\\ \hline
$D_{\phi,\theta}$ 				& 1					& $\bar{J}_{1,2}$	    & 0.82 \si{Kg\cdot m^2}\\ \hline
$\ddot{\mathbf{X}}$	Limit		& \si{\pm 3\ m/s^2}	& $\bar{J}_3$ 	    	& 1.49 \si{Kg\cdot m^2}\\ \hline
$\max|\delta_j|$				& $0.12$	    	& $\max|J_{\Delta,j}|$ 	& $0.3$\\ \hline
$\max|K_{\Delta,j}|$		 	& $0.1$		    	& $\zeta$ 				& 0.707\\ \hline
\end{tabular}
\end{table}

\begin{figure}[t]
\begin{center}
\includegraphics[width=8.5cm]{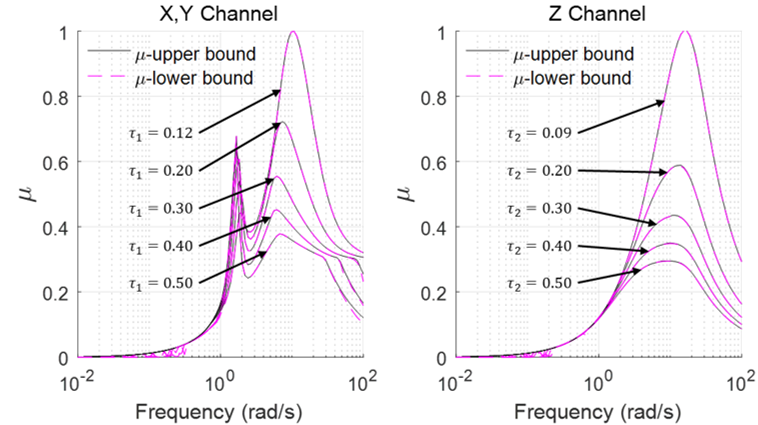}
\end{center}
\vspace{-0.5cm}
\caption{$\mu$-analysis results for $X$, $Y$ channel (left), and $Z$ channel (right).}
\label{gp:mu_result}
\end{figure}

Table \ref{table_1} shows the multi-rotor's physical quantities and controller gains used both in  the simulation and the experiment.
The gains $P_{\phi, \theta}$ and $D_{\phi, \theta}$ are predefined values set during the primary gain-tuning process to obtain the ability to control the attitude of the platform.
The translational acceleration limit is set to prevent flight failure due to excessive acceleration control inputs and is set at $\pm 3\ \si{m/s^2}$ to have a roll and pitch limit of approximately $\pm0.3\ \si{rad}$ in level flight condition.
As previously mentioned, the unmodeled time delay $\delta_j$ is set to 0.1, and the gain error is assumed to be a maximum error of 10 percent.
For MOIs that are difficult to estimate, we assumed a wider 30 percent uncertainty.
The damping ratio $\zeta$ of the second order filter is set to 0.707, which is the critical damping ratio, to balance the overshoot and late response.
Fig. \ref{gp:mu_result} shows the results of $\mu$-analysis. 
From the analysis, we can see that the system is stable when $\tau_1>0.12$ and $\tau_2>0.09$.

\begin{figure}[t]
\begin{center}
\includegraphics[width=8.5cm]{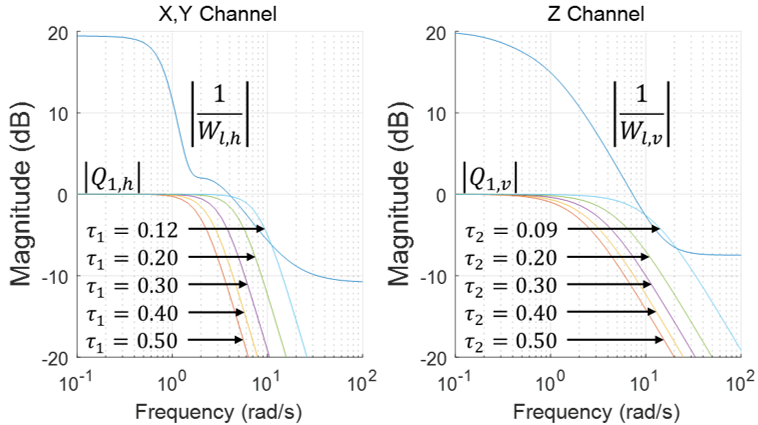}
\end{center}
\vspace{-0.5cm}
\caption{SGT-based analysis results for $X$, $Y$ channel (left), and $Z$ channel (right).}
\label{gp:small_gain}
\end{figure}

Fig. \ref{gp:small_gain} shows the results of the SGT-based stability analysis, performed in the same manner as \cite{DOB_sjlazza}. 
The analysis is based on the following model:
\begin{equation}
\label{eq:plant_SGT}
\begin{split}
{P}_j(s)={\it{\Lambda_{n,n,j}}}(s)(1+{\Delta}_{l,j}{\it{{W}}}_{l,j}),\ ||{\Delta}_{l,j}||_\infty\leq1,
\end{split}
\end{equation}
where all uncertainties due to $\delta_j$, $K_j$ and $J_j$ are lumped using the functions  $W_{l,h} (=W_{l,1}, W_{l,2})$ and $W_{l,v} (=W_{l,3})$, whose magnitude increases over frequency as shown in blue curves of Fig. \ref{gp:small_gain}. The stability condition of the SGT-based analysis in this case is
\begin{equation}
\label{eq:SGT_condition}
\bar{\sigma}\big({Q}_j(j\omega)\big)\bar{\sigma}\big({\it{W}}_{l,j}(j\omega)\big)<1
\end{equation}
\cite{DOB_sjlazza,SGTnmu,Small_Gain_Theorem}. In the SGT-based analysis, the bode plots of the $Q$-filter with $\tau_1=0.12$ and $\tau_2=0.09$ indicate that system with those $\tau$ values could be unstable. However, through the $\mu$-analysis, those $\tau$ values are still in the stable region. From this, we can confirm that the $\mu$-analysis provides more rigorous $\tau$ boundary values than SGT-based analysis.

\section{SIMULATION AND EXPERIMENTAL RESULT}
This section reports simulation and experimental results to validate the performance of our three-dimensional force controller and the disturbance cancellation performance of the DOB technique. 
The comparison of the acceleration tracking performance of the force control methods according to the MOI variation is already shown in the simulation of Fig. \ref{gp:pos_acc_comparison}.
Therefore, in this section, we provide  
\begin{enumerate}
    \item experimental result to demonstrate the performance of the proposed force control technique for the actual plant, and
    \item simulation and experimental results to demonstrate the capability of the DOB in overcoming the translational force disturbance.
\end{enumerate}
Based on the results from the previous section, the cutoff frequencies of the $Q$-filter are set to $\tau_1=0.15$ and $\tau_2=0.12$ in both simulation and actual experiment with additional margins to ensure additional stability. 

\subsection{Validation of acceleration tracking performance}

\begin{figure}[t]
\begin{center}
\includegraphics[width=8.5cm]{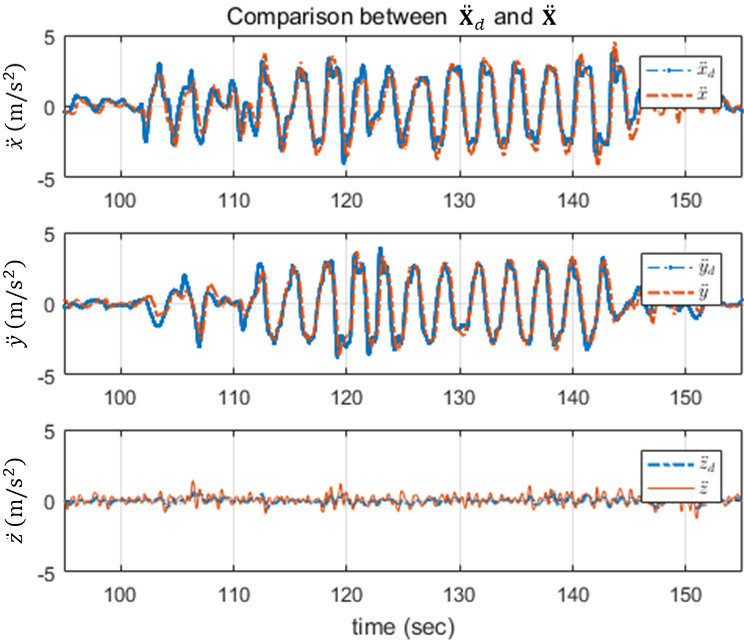}
\end{center}
\vspace{-0.5cm}
\caption{[Experiment] Desired 3-D acceleration generated by the operator through the R/C controller (blue), and the actual acceleration (red dash) generated by multi-rotor.}
\label{gp:Xd2X}
\end{figure}

In the experiment, arbitrary desired acceleration commands for $X$ and $Y$ directions are given by the operator-controlled radio controller. Fig. \ref{gp:Xd2X} shows the multi-rotor accurately following the target acceleration. From this result, we can confirm that our three-dimensional translational acceleration control technique functions effectively even in the actual flight.

\subsection{Validation of DOB performance}
\subsubsection{Simulation result}
\begin{figure}[t]
\begin{center}
\includegraphics[width=8.5cm]{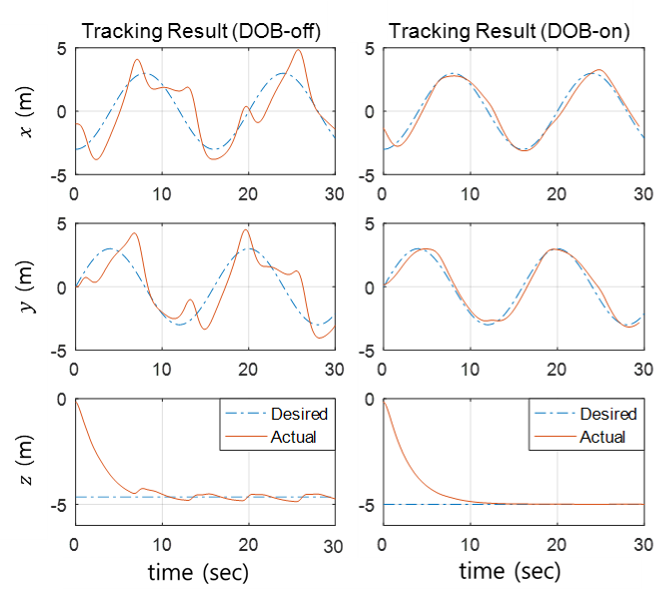}
\end{center}
\vspace{-0.5cm}
\caption{[Simulation] Comparison of trajectory tracking performance before (left) and after (right) applying the DOB algorithm.}
\label{gp:sim_position}
\end{figure}
In the simulation, the multi-rotor follows a circular trajectory with radius of 3 \si{m} and height of 5 \si{m}. Meanwhile, the multi-rotor is exposed to periodic disturbances with accelerations up to 5.5 \si{m/s^2} in each axis. Fig. \ref{gp:sim_position} compares the multi-rotor's position tracking performance before and after applying DOB. On the left graphs of Fig. \ref{gp:sim_position}, the target trajectory tracking results are not smooth due to the unexpected disturbances, whereas the trajectory deviation is drastically reduced in the right graphs where the DOB algorithm is applied. 

\subsubsection{Experimental Result}
\begin{figure}[t]
\begin{center}
\includegraphics[width=8.72cm]{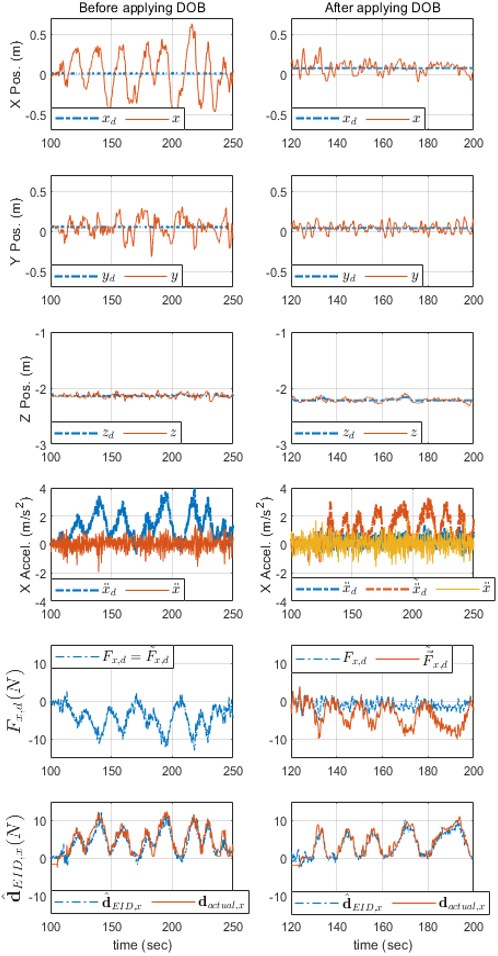}
\end{center}
\vspace{-0.5cm}
\caption{Comparison of the target position tracking performance before (left) and after (right) the DOB algorithm is applied.}
\label{gp:exp_position_force}
\end{figure}
\begin{figure}[t]
\begin{center}
\includegraphics[width=8.8cm]{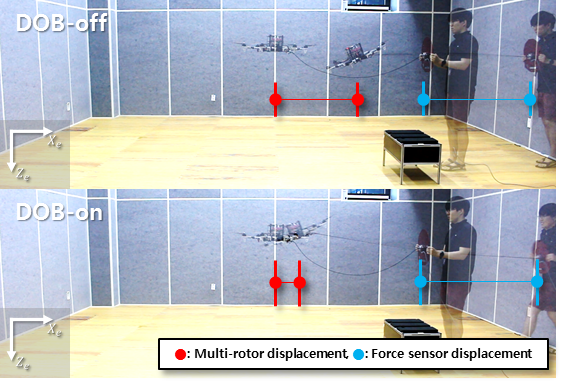}
\end{center}
\vspace{-0.5cm}
\caption{Experiment for DOB performance validation with disturbance using a tether.  A force sensor is attached to the tether only to check the disturbance estimation performance. }
\label{gp:experiment}
\end{figure}

In the experiment, the multi-rotor is commanded to hover at a specific point in three-dimensional space but connected to the translational force measurement sensor via the tether to measure the applied disturbance force. As we can see in Fig. \ref{gp:experiment}, the operator aligns the force sensor in the $X$-axis and pulls and releases the force sensor periodically to apply a disturbance to the multi-rotor. 

Fig. \ref{gp:exp_position_force} is a comparison of hovering performance before (left) and after (right) applying the DOB algorithm. The graphs in the left column are the case when the DOB is not applied, which has a larger $X$ directional position shift than other axes. Unlike the DOB-off case, the DOB-on case shows a significant reduction in position error. 
\begin{figure}[t]
\begin{center}
\includegraphics[width=8.8cm]{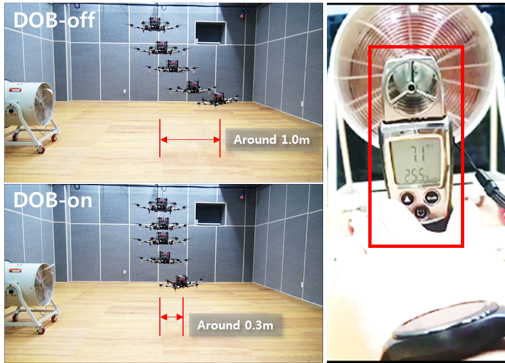}
\end{center}
\vspace{-0.4cm}
\caption{Comparison of the target position tracking performance in wind blast environment using  an industrial fan.}
\label{gp:wind_DOB}
\end{figure}
Two graphs at the forth row shows the acceleration tracking results.
When DOB is not applied, an acceleration signal is generated by the position error, but we can see that the target acceleration cannot be followed due to the disturbance.
Meanwhile, we can see that the acceleration of the platform (yellow solid line) well tracks the target acceleration (blue dash-single dotted line).
This is because the well-behaved DOB algorithm generated control input including the disturbance compensation signal (orange dash-single dotted line) and applied to the platform.
The effect of the DOB can be confirmed by significantly reduced position error.
Four graphs at the bottom of the figure show the difference between the signal $\vec{\mathcal{F}}_d$ and $\tilde{\vec{\mathcal{F}}}_d$ (fifth row), and the comparison between $d_{actual,x}$ measured by force sensor and $\hat{d}_{EID,x}$ estimated by DOB algorithm (sixth row). When DOB is not applied, $\hat{d}_{EID}$ estimation process is working internally but the signal is not merged into $\tilde{\vec{\mathcal{F}}}_d$ signal, making $\vec{\mathcal{F}}_d$ and $\tilde{\vec{\mathcal{F}}}_d$ have the same value. On the other hand, we can see the difference between the $\vec{\mathcal{F}}_d$ and the $\tilde{\vec{\mathcal{F}}}_d$ signal when DOB is applied, because the $\hat{d}_{EID}$ signal is merged into the $\tilde{\vec{\mathcal{F}}}_d$ signal. Two graphs in the last row show the comparison between the measured disturbance and the estimated disturbance, and we can confirm that the estimates are fairly accurate in both cases.

An extra flight experiment is conducted under wind disturbance to validate the DOB performance in a more realistic  environment. As we can see in Fig. \ref{gp:wind_DOB}, the target location of the multi-rotor is set on the centerline of a strong wind generator that generates wind speed of 7 \si{m/s}. The performance of DOB is visualized by comparing the position difference between DOB-on and DOB-off situations.  the multi-rotor has a position error of about 1 \si{m} in the DOB-off case and about 0.3 \si{m} in the DOB-on case. Through the experiment, we can confirm that the proposed DOB algorithm works effectively even against a wind disturbance. 

\section{CONCLUSION}

In this paper, we introduced 1) a new method of converting the target acceleration command to the desired attitude and total thrust, and 2) a DOB method for overcoming the disturbance that obstructs the translational motion, to more accurately control the translational acceleration of the multi-rotor UAV. 
In the control input conversion process, we reflect the different dynamic characteristics of attitude and thrust, so that more precise control is possible than the existing methods. 
Then, by using the DOB-based robust control algorithm based on the nominal translational force system, the magnitude of the disturbance force applied to the fuselage is estimated and compensated. 
For the robust stability guarantee, the $Q$-filter of the DOB is designed based on the $\mu$-stability analysis. 
The validity of the proposed method is confirmed through simulation and actual experiments. 

The proposed technique is useful in various applications such as aerial parcel delivery service or drone-based industrial operations where precise acceleration control is required. For example, 
in a multi-rotor-based parcel delivery service, the proposed DOB algorithm can maintain the nominal flight performance by considering the additional force due to the weight of the cargo attached to the multi-rotor as a disturbance to be estimated. Also, the proposed algorithm is suitable for situations that require precise trajectory tracking performance even in  windy conditions such as maritime operations or human-rescue missions.
For industrial applications involving collaborative flight of multiple multi-rotors, the proposed algorithm can be used to estimate and stabilize internal forces caused in between physically-coupled multi-rotors.

\appendices




\ifCLASSOPTIONcaptionsoff
  \newpage
\fi



\bibliographystyle{Bibliography/IEEEtran}
\bibliography{Bibliography/IEEEabrv,Bibliography/BIB_1x-TIE-2xxx}\ 

\begin{thebibliography}{10}
\providecommand{\url}[1]{#1}
\csname url@samestyle\endcsname
\providecommand{\newblock}{\relax}
\providecommand{\bibinfo}[2]{#2}
\providecommand{\BIBentrySTDinterwordspacing}{\spaceskip=0pt\relax}
\providecommand{\BIBentryALTinterwordstretchfactor}{4}
\providecommand{\BIBentryALTinterwordspacing}{\spaceskip=\fontdimen2\font plus
\BIBentryALTinterwordstretchfactor\fontdimen3\font minus
  \fontdimen4\font\relax}
\providecommand{\BIBforeignlanguage}[2]{{%
\expandafter\ifx\csname l@#1\endcsname\relax
\typeout{** WARNING: IEEEtran.bst: No hyphenation pattern has been}%
\typeout{** loaded for the language `#1'. Using the pattern for}%
\typeout{** the default language instead.}%
\else
\language=\csname l@#1\endcsname
\fi
#2}}
\providecommand{\BIBdecl}{\relax}
\BIBdecl

\bibitem{mUAVDynamics}
R.~Beard, ``Quadrotor dynamics and control rev 0.1,'' 2008.

\bibitem{mofid2018adaptive}
O.~Mofid and S.~Mobayen, ``Adaptive sliding mode control for finite-time
  stability of quad-rotor uavs with parametric uncertainties,'' \emph{ISA
  transactions}, vol.~72, pp. 1--14, 2018.

\bibitem{zuo2010trajectory}
Z.~Zuo, ``Trajectory tracking control design with command-filtered compensation
  for a quadrotor,'' \emph{IET control theory \& applications}, vol.~4, no.~11,
  pp. 2343--2355, 2010.

\bibitem{zhao2015nonlinear}
B.~Zhao, B.~Xian, Y.~Zhang, and X.~Zhang, ``Nonlinear robust adaptive tracking
  control of a quadrotor uav via immersion and invariance methodology,''
  \emph{IEEE Transactions on Industrial Electronics}, vol.~62, no.~5, pp.
  2891--2902, 2015.

\bibitem{mobayen2016lmi}
S.~Mobayen and F.~Tchier, ``An lmi approach to adaptive robust tracker design
  for uncertain nonlinear systems with time-delays and input nonlinearities,''
  \emph{Nonlinear Dynamics}, vol.~85, no.~3, pp. 1965--1978, 2016.

\bibitem{mobayen2018composite}
------, ``Composite nonlinear feedback integral sliding mode tracker design for
  uncertain switched systems with input saturation,'' \emph{Communications in
  Nonlinear Science and Numerical Simulation}, vol.~65, pp. 173--184, 2018.

\bibitem{ohnishi1996motion}
K.~Ohnishi, M.~Shibata, and T.~Murakami, ``Motion control for advanced
  mechatronics,'' \emph{IEEE/ASME transactions on mechatronics}, vol.~1, no.~1,
  pp. 56--67, 1996.

\bibitem{Attitude_DOB_1}
W.~Qingtong, W.~Honglin, W.~Qingxian, and C.~Mou, ``Backstepping-based attitude
  control for a quadrotor uav using nonlinear disturbance observer,'' in
  \emph{34th Chinese Control Conference (CCC)}.\hskip 1em plus 0.5em minus
  0.4em\relax IEEE, 2015, pp. 771--776.

\bibitem{Attitude_DOB_2}
A.~T. Salton, D.~Eckhard, J.~V. Flores, G.~Fernandes, and G.~Azevedo,
  ``Disturbance observer and nonlinear damping control for fast tracking
  quadrotor vehicles,'' in \emph{IEEE Conference on Control Applications
  (CCA)}.\hskip 1em plus 0.5em minus 0.4em\relax IEEE, 2016, pp. 705--710.

\bibitem{Attitude_DOB_3}
T.~Tomic, ``Evaluation of acceleration-based disturbance observation for
  multicopter control,'' in \emph{European Control Conference (ECC)}.\hskip 1em
  plus 0.5em minus 0.4em\relax IEEE, 2014, pp. 2937--2944.

\bibitem{Attitude_DOB_4}
K.~Lee, J.~Back, and I.~Choy, ``Nonlinear disturbance observer based robust
  attitude tracking controller for quadrotor uavs,'' \emph{International
  Journal of Control, Automation and Systems}, vol.~12, no.~6, pp. 1266--1275,
  2014.

\bibitem{Attitude_DOB_5}
H.~Wang and M.~Chen, ``Trajectory tracking control for an indoor quadrotor uav
  based on the disturbance observer,'' \emph{Transactions of the Institute of
  Measurement and Control}, vol.~38, no.~6, pp. 675--692, 2016.

\bibitem{General_DOB}
A.~Aboudonia, R.~Rashad, and A.~El-Badawy, ``Time domain disturbance observer
  based control of a quadrotor unmanned aerial vehicle,'' in \emph{XXV
  International Conference on Information, Communication and Automation
  Technologies (ICAT)}.\hskip 1em plus 0.5em minus 0.4em\relax IEEE, 2015, pp.
  1--6.

\bibitem{neural_DOB}
P.~Castaldi, N.~Mimmo, R.~Naldi, and L.~Marconi, ``Robust trajectory tracking
  for underactuated vtol aerial vehicles: Extended for adaptive disturbance
  compensation,'' \emph{IFAC Proceedings Volumes}, vol.~47, no.~3, pp.
  3184--3189, 2014.

\bibitem{suseong_DOB}
S.~Kim, S.~Choi, H.~Kim, J.~Shin, H.~Shim, and H.~J. Kim, ``Robust control of
  an equipment-added multirotor using disturbance observer,'' \emph{IEEE
  Transactions on Control Systems Technology}, vol.~26, no.~4, 2018.

\bibitem{DOB_star_1}
A.~Chovancov{\'a}, T.~Fico, P.~Hubinsk{\`y}, and F.~Ducho{\v{n}}, ``Comparison
  of various quaternion-based control methods applied to quadrotor with
  disturbance observer and position estimator,'' \emph{Robotics and Autonomous
  Systems}, vol.~79, pp. 87--98, 2016.

\bibitem{DOB_star_2}
W.~Dong, G.-Y. Gu, X.~Zhu, and H.~Ding, ``High-performance trajectory tracking
  control of a quadrotor with disturbance observer,'' \emph{Sensors and
  Actuators A: Physical}, vol. 211, pp. 67--77, 2014.

\bibitem{DOB_sjlazza}
S.~J. Lee, S.~Kim, K.~H. Johansson, and H.~J. Kim, ``Robust acceleration
  control of a hexarotor uav with a disturbance observer,'' in \emph{IEEE 55th
  Conference on Decision and Control (CDC)}.\hskip 1em plus 0.5em minus
  0.4em\relax IEEE, 2016, pp. 4166--4171.

\bibitem{proof_of_small_gyroscopic}
H.~Lee and H.~J. Kim, ``Trajectory tracking control of multirotors from
  modelling to experiments: A survey,'' \emph{International Journal of Control,
  Automation and Systems}, vol.~15, no.~1, pp. 281--292, 2017.

\bibitem{EID}
J.-H. She, M.~Fang, Y.~Ohyama, H.~Hashimoto, and M.~Wu, ``Improving
  disturbance-rejection performance based on an equivalent-input-disturbance
  approach,'' \emph{IEEE Transactions on Industrial Electronics}, vol.~55,
  no.~1, pp. 380--389, 2008.

\bibitem{DOB_Q}
C.~J. Kempf and S.~Kobayashi, ``Disturbance observer and feedforward design for
  a high-speed direct-drive positioning table,'' \emph{IEEE Transactions on
  control systems Technology}, vol.~7, no.~5, pp. 513--526, 1999.

\bibitem{SGTnmu}
C.~Fielding, A.~Varga, S.~Bennani, and M.~Selier, \emph{Advanced techniques for
  clearance of flight control laws}.\hskip 1em plus 0.5em minus 0.4em\relax
  Springer Science \& Business Media, 2002, vol. 283.

\bibitem{sariyildiz2013analysis}
E.~Sariyildiz and K.~Ohnishi, ``Analysis the robustness of control systems
  based on disturbance observer,'' \emph{International Journal of Control},
  vol.~86, no.~10, pp. 1733--1743, 2013.

\bibitem{kim2018robust}
S.~Kim, J.~Park, S.~Kang, P.~Y. Kim, and H.~J. Kim, ``A robust control approach
  for hydraulic excavators using $\mu$-synthesis,'' \emph{International Journal
  of Control, Automation and Systems}, vol.~16, no.~4, pp. 1615--1628, 2018.

\bibitem{muDoyle}
J.~Doyle, ``Analysis of feedback systems with structured uncertainties,'' in
  \emph{IEE Proceedings D-Control Theory and Applications}, vol. 129,
  no.~6.\hskip 1em plus 0.5em minus 0.4em\relax IET, 1982, pp. 242--250.

\bibitem{mu-analysis}
K.~Zhou and J.~C. Doyle, \emph{Essentials of robust control}.\hskip 1em plus
  0.5em minus 0.4em\relax Prentice hall Upper Saddle River, NJ, 1998, vol. 104.

\bibitem{Small_Gain_Theorem}
S.~Skogestad and I.~Postlethwaite, \emph{Multivariable feedback control:
  analysis and design}.\hskip 1em plus 0.5em minus 0.4em\relax Wiley New York,
  2007, vol.~2.

\end{thebibliography}
%



%

\begin{IEEEbiography}[{\includegraphics[width=1in,height=1.25in,clip,keepaspectratio]{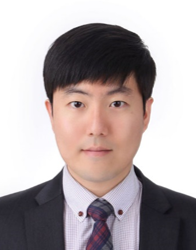}}]
{Seung Jae Lee} (S'17) received the B.S. degree in mechanical engineering from Hanyang University, Seoul, Korea in 2014, and the M.S. degree in mechanical and aerospace engineering from Seoul National University, Seoul, Korea in 2016. 

He is currently a Ph.D. candidate in mechanical and aerospace engineering at the Seoul National University, Seoul, Korea. His research interests include robust control theory, new robot design, system identification and state estimation.
\end{IEEEbiography}

\vspace{-0cm}
\begin{IEEEbiography}[{\includegraphics[width=1in,height=1.25in,clip,keepaspectratio]{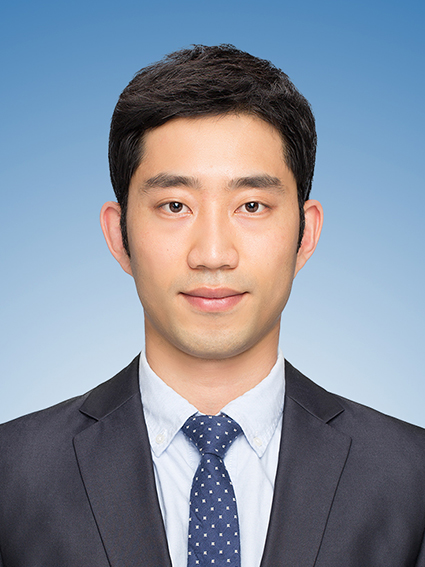}}]
{Seunghyun Kim} (S'12) received the B.S. degree in mechanical engineering from Hanyang University, Seoul, Korea, and the M.S. and Ph.D. degrees in mechanical and aerospace engineering from Seoul National University, Seoul, Korea. 

He is currently Research Engineer with Hyundai Motor Company, Hwasung, Korea. His research interests include robust control nonlinear control.

\end{IEEEbiography}

\vspace{-0cm}
\begin{IEEEbiography}[{\includegraphics[width=1in,height=1.25in,clip,keepaspectratio]{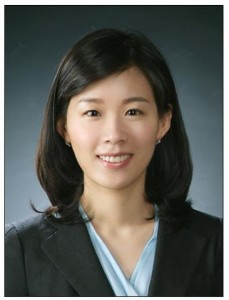}}]
{Hyoun Jin Kim} (S'98-M'02) received the B.S. degree from the Korea Advanced Institude of Technology, Daejeon, South Korea, in 1995, and the M.S. and Ph.D. degrees in mechanical engineering from the University of California at Berkeley (UC Berkeley), Berkeley, CA, USA, in 1999 and 2001, respectively.

From 2002 to 2004, she was a Post-Doctoral Researcher in electrical engineering and computer science with UC Berkeley. In 2004, she joined the Department of Mechanical and Aerospace Engineering, Seoul National University, Seoul, South Korea, as an Assistant Professor, where she is currently a Professor. Her current research interests include intelligent control of robotic systems and motion planning.
\end{IEEEbiography}




\end{document}